\documentclass[letterpaper]{article} 
\usepackage[]{aaai24}  
\usepackage{times}  
\usepackage{helvet}  
\usepackage{courier}  
\usepackage[hyphens]{url}  
\usepackage{graphicx} 
\usepackage{natbib}  
\usepackage{caption} 
\usepackage{algorithm}
\usepackage{algorithmic}

\usepackage{newfloat}
\usepackage{listings}
\DeclareCaptionStyle{ruled}{labelfont=normalfont,labelsep=colon,strut=off} 
\floatstyle{ruled}
\newfloat{listing}{tb}{lst}{}
\floatname{listing}{Listing}
\title{Distilled ChatGPT Topic \& Sentiment Modeling with Applications in Finance}
\author {
    Olivier Gandouet\equalcontrib\textsuperscript{\rm 1},
    Mouloud Belbahri\equalcontrib\textsuperscript{\rm 2},
    Armelle Jezequel\textsuperscript{\rm 1},
    Yuriy Bodjov \textsuperscript{\rm 1}
}
\affiliations {
    \textsuperscript{\rm 1}
    TD Asset Management, 
    \textsuperscript{\rm 2}Layer 6 AI\\
    olivier.gandouet@tdam.com, mouloud@layer6.ai, 
    armelle.jezequel@tdam.com, yuriy.bodjov@tdam.com
}

\begin{document}

\maketitle

\begin{abstract}


In this study, ChatGPT is utilized to create streamlined models that generate easily interpretable features. These features are then used to evaluate financial outcomes from earnings calls. We detail a training approach that merges knowledge distillation and transfer learning, resulting in lightweight topic and sentiment classification models without significant loss in accuracy. These models are assessed through a dataset annotated by experts. The paper also delves into two practical case studies, highlighting how the generated features can be effectively utilized in quantitative investing scenarios.




\end{abstract}


\section{Introduction}

Quarterly earnings conference calls serve as a valuable source of information for investors and analysts, offering insights into a company's financial performance, strategic initiatives, and overall business outlook. These calls involve key executives discussing financial results, market trends, and potential opportunities or challenges. Investors and analysts rely on these calls to understand a company's financial health, growth prospects, and future plans \cite{arslan2016managers}. However, the complexity and extent of these calls can make it challenging for stakeholders to focus on specific subjects of interest, requiring additional time and effort for analysis. Moreover, analyzing hundreds or thousands of earnings calls simultaneously is labor-intensive. Currently, data providers offer quarterly earning call transcripts, which serve as the foundation for building and training models to understand and categorize the key messages within the transcripts. The investment community has shown a strong interest in developing models that can extract useful features from earnings call transcripts for forecasting stock returns, assessing risk, or improving the analysis efficiency of these calls.

Techniques such as Topic Modeling, NLP, and Data Mining, employed in machine learning, have proven successful in extracting key themes and sentiments from text across a broad range of fields \cite{hajek2013evaluating}.
Topic Modeling can help identify key themes discussed during earnings calls, allowing for a deeper understanding of factors driving returns. However, traditional Topic Modeling techniques, such as Latent Dirichlet Allocation \cite{blei2003latent}, have limitations such as fixed number of topics and the need for extensive pre-processing \cite{zhao2021topic}. Implementing Topic Modeling in the financial domain requires expertise and can be sensitive to noise, emphasizing the need for more advanced approaches to enhance its effectiveness and reliability.
The advent of Large Language Models (LLMs) offers a chance to address certain limitations of conventional topic models, especially in quantitative equity research. Here, powerful LLMs, having processed numerous data points, possess strong priors that allow for efficient analysis of the quaterly earnings calls.
However, the adoption of these sophisticated models yield specific limitations, such as high computational demands, reproducibility,  explainability, and reliability issues (including instances of model hallucination).

In the area of investment management research, using heavy LLMs raises concerns about lack of control over model outputs, computational constraints, necessitating the need for close monitoring and control throughout the process. To address these challenges, we propose a hybrid approach that combines LLMs with human supervision for more efficient and precise data annotation. Our contribution includes a knowledge distillation method (KD) that reduces the computational demand of LLMs, allowing for the implementation of the framework with fewer resources. In the following sections, we will demonstrate the proposed methodology using an example, outlining the steps involved in building an earnings call topic/sentiment model at a lower cost.

\section{Knowledge Distillation Pipeline}

Let us consider a corpus containing a collection of documents $\mathcal{D}$. Any document $d_i \in \mathcal{D}$ is composed of $J_i$ sentences and its $j$th sentence $s_{ij}$ consists of $w_{ij}$ words. We further denote by $\mathcal{S}$ the collection of sentences in all the documents and by $\mathcal{K}$ the collection of topics in all the sentences. With these notations, the task for a topic model is to learn the latent topics of $\mathcal{K}$ from the observed sentences $\mathcal{S}$.

\begin{figure}
     \centering
     \includegraphics[width=0.45\textwidth]{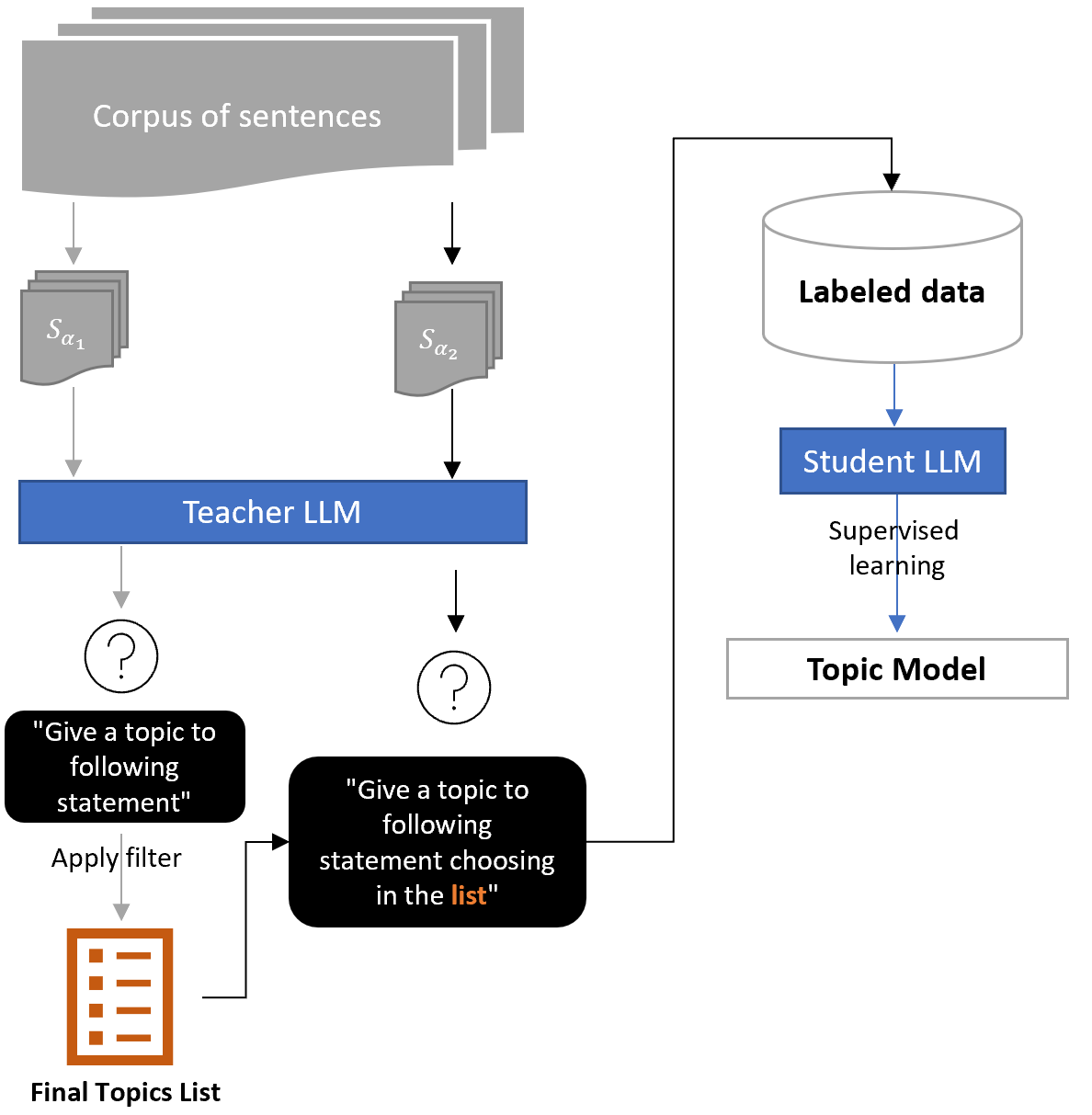}
     \caption{Earning Calls Topic Classification Pipeline.}
     \label{fig:Pipeline}
\end{figure}

The first objective of our study is to construct a topic classification model. To achieve this, we present a three-step methodology that capitalizes on the inherent language knowledge possessed by contemporary LLMs in the finance domain. Our approach aims to transfer this understanding to a smaller-scale model, while maintaining a certain level of accuracy comparable to larger models. The complete pipeline is illustrated in Figure~\ref{fig:Pipeline}. The three steps are:
\begin{enumerate}
    \item Identifying a comprehensive list of topics that adequately represent a significant portion of the subject matter in the field.
    \item Creating a labeled dataset of sentences from the corpus based on the teacher topic model.
    \item Training a ``small" topic model using a supervised approach.
\end{enumerate}

\subsubsection{Earning Calls Data}
We have access to earning call transcripts obtained from Factset, covering the years 2010 to 2023. 
The number of distinct sentences to consider in these transcripts is extensive, exceeding 60 million sentences. However, it is widely recognized that the number of relevant general topics in the finance domain is considerably smaller.

\subsubsection{Identifying Financial Topics}

The first step consists of identifying a comprehensive list of financial topics from sentences of publicly available conference calls transcripts. To do so, we consider only a (small) random  proportion $\alpha_1 << 1$ of all the sentences $\mathcal{S}$ in the corpus. We denote this sample by $\mathcal{S}_{\alpha_1}$ which is then used as the input data for ChatGPT to get an answer to the following question.

\begin{center}
\textbf{``Can you provide financial topics that would describe the following sentences using a general classification?"}    
\end{center}

Next, we incorporated the following prompt into the process.

\begin{center}
\textbf{``Format your answer by separating all the detected topics with semi-colons."}
\end{center}

\begin{figure}
    \centering
    \includegraphics[width=0.47\textwidth]{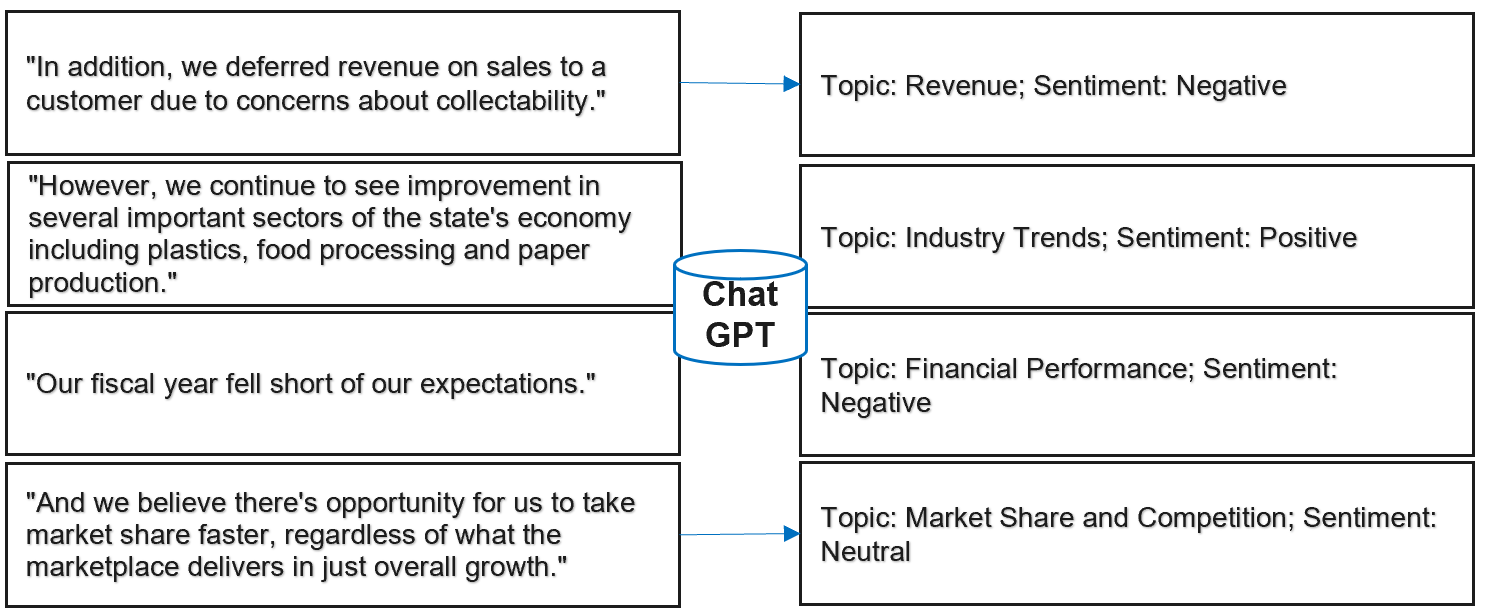}
    \caption{Examples of sentences labeled by Chat GPT.}
    \label{fig:label-dataset}
\end{figure}

Examples of sentences to label and suggested topics can be found in Figure~\ref{fig:label-dataset}. From this first step, we obtained about 100 topics suggested by ChatGPT.

In order to proceed with a classification model, it is better to analyze and potentially decrease the number of topics. There are various alternative methods to accomplish this. One approach that can be used is the ``LLM approach". This involves presenting the list of output topics to the teacher LLM and requesting it to reduce the size by selecting only the non-redundant and meaningful topics for companies' earnings calls. This method is sufficient to obtain a revised and smaller set of topics. Another way to accomplish this task is by filtering topics based on a pre-defined threshold. Let $K$ be the number of suggested topics and $n_k$ be the number of sentences from the sample $\mathcal{S}_{\alpha_1}$ that were classified as the $k$th topic, $k=1, \ldots, K$. The idea is to keep the most frequent topics (in terms of $n_k$) allowing the classification of a total number of sentences greater than the pre-defined threshold (we used 2\% in our use case) .
A third method to decrease the suggested topics is by using sentence embedding. This approach involves employing a pre-trained sentence embedding model to convert the list of outputted topics into vectors. The vectors can then be fed to the $k$-means algorithm, which clusters the topics, resulting in a reduced final number of topics. Finally, an alternative approach is to seek expert opinion. Given the relatively small number of proposed topics, it is feasible to obtain input from experts to further refine the list. This can be accomplished by requesting a smaller subset of sentences for their evaluation.

\begin{figure}[!ht]
    \centering
    \includegraphics[width=0.47\textwidth]{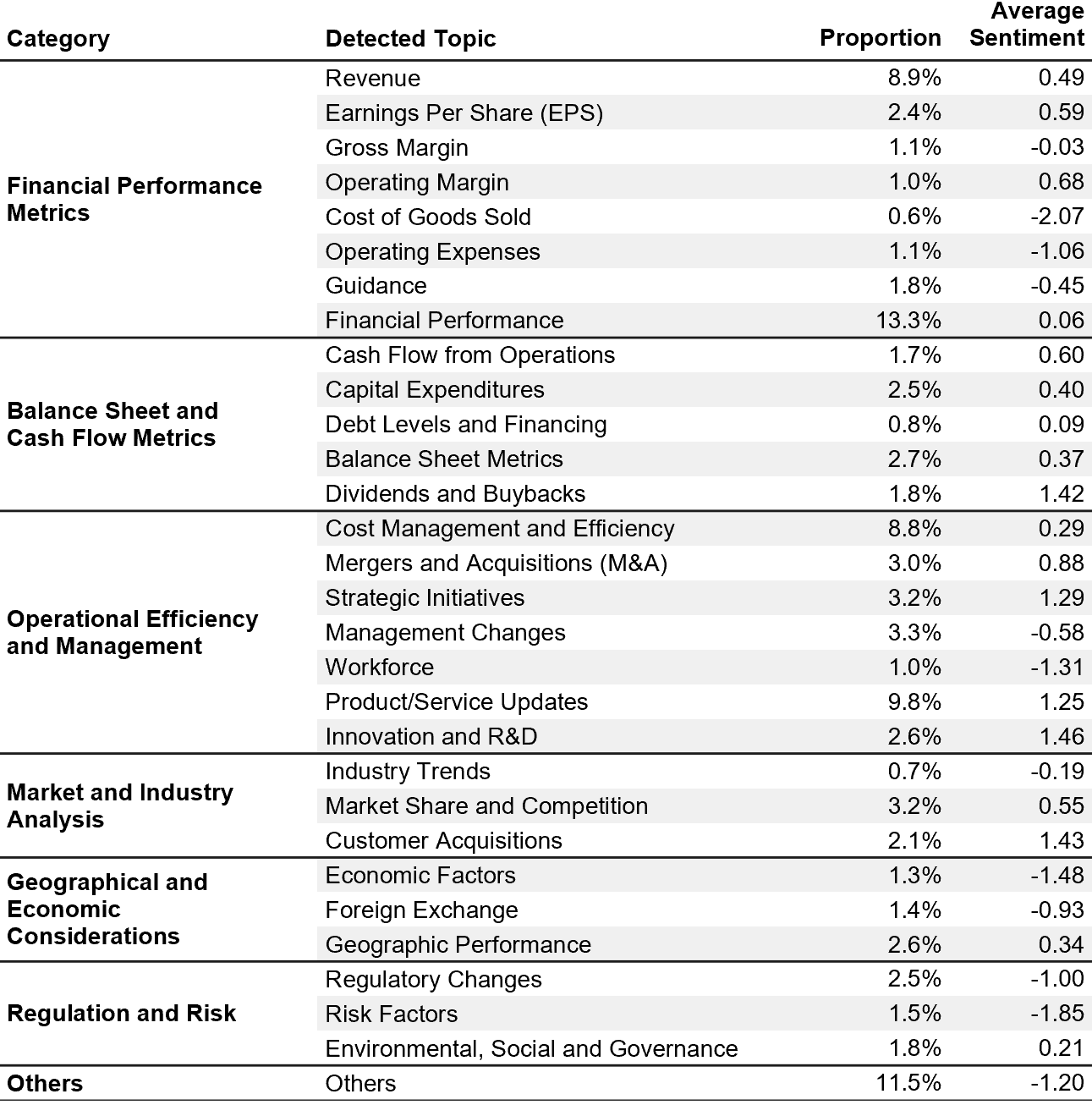}
    \caption{Identified topics distribution and average sentiment per topic on the labeled sentences dataset.}
    \label{fig:statistics_topics}
\end{figure}

\subsubsection{Creating a Labeled Dataset of Sentences}

In the second step, we take a new random sample of sentences $\mathcal{S}_{\alpha_2}$. In our application, we randomly sampled 80,000 sentences from the earning calls data. We then asked the following question using the final reduced list of topics from the first step. 

\begin{center}
\textbf{``Considering the following list of topics: [\textit{list of topics}], could you provide a classification on the following sentences topics? Please format your answer in the following way: Topic: Topic identified."}
\end{center}

Figure~\ref{fig:statistics_topics} presents descriptive statistics for the final list of topics. It should be noted that the responses generated did not fully comply with the specified output format, likely due to the commonly recognized issue of model hallucinations. To improve the quality of these outputs, we used regular expressions to verify their adherence to the desired format and discarded those that did not match. This filtering procedure yielded a refined labeled database of around 50,000 sentences. This curated dataset is now suitable for training our refined topic classification model.

\subsection{Training a Topic Classification Model}

\begin{figure}[!ht]
     \centering
     \includegraphics[width=0.25\textwidth]{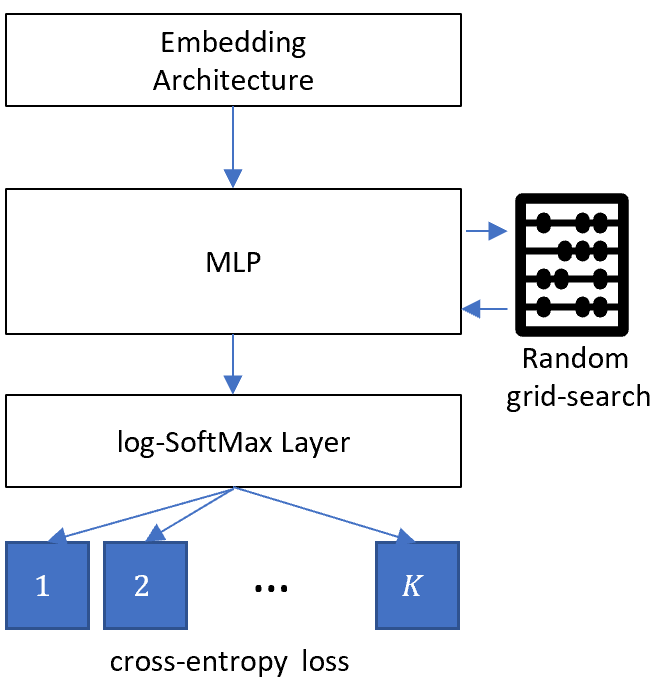}
     \caption{Topic Classification Student Model Architecture.}
     \label{fig:topic-model-architecture}
\end{figure}

To create a lightweight version topic classification model, we chose to use a lighter model architecture such as MPNet~\cite{song2020mpnet}, a language model built upon the popular BERT architecture. 
We added an MLP to the output of the MPNet model as shown in Figure~\ref{fig:topic-model-architecture}. We compared different pre-trained models to MPNet and report the results in Table~\ref{tab:topic_perf}.

The MLP architecture was determined through a random search within the parameter space defined in Table~\ref{tab:hyper_search}. For each layer, we applied the ReLU activation. To control for overfitting, we made use of dropout at different rates. Furthermore, considering the non-linearity of ReLU, we incorporated batch normalization in all layers to account for this effect. To establish a search space that promotes convergence, we conducted initial tests on our dataset, specifically investigating the batch size and the learning rate. We performed $50$ trials on a reduced training dataset, which consisted of $60\%$ of the original training data. We further divided it into a sub-training set ($80\%$) and a sub-validation set ($20\%$), resulting in approximately $24,000$ sentences for training and $6,000$ for validation. To control for the number of epochs, we used early stopping. We determined the learning parameters that yielded the highest $F_1$ score on the remaining $20,000$ validation sentences. The final model was retrained using the complete original training set, which was divided into an $80\%-20\%$ split for training and validation.

\begin{table}[!ht]
    \centering
    \resizebox{0.47\textwidth}{!}{\begin{tabular}{l|c}
    \hline
    \bf Parameter & \bf Search Space\\
    \hline
    Hidden Layers &	$\{1,2,3,4\}$ \\ 
    First Layer Size	& $\{1024,768,512,256,128\}$ \\
    Dropout Rate & $\{0,0.4,0.7\}$ \\
    With Batch Norm & \{True, False\} \\
    Layer Ratio & $\{0.5,1\}$\\
    Learning Rate & $\{0.00005,0.0001,0.0002,0.0004,0.0008,0.001\}$\\
    Batch Size & $\{64,128,256,512\}$\\
    \hline
    \end{tabular}}
    \caption{Topic Classification Hyperparameters Space.}
    \label{tab:hyper_search}
\end{table}

\begin{table}[]
    \centering
    \resizebox{0.47\textwidth}{!}{\begin{tabular}{l|c|c|c|c}
    \hline
     \bf Model    & \bf \#Tokens & \bf Size	& \bf $F_1$ vs. Teacher &	\bf $F_1$ vs. Human\\
    \hline
    Paraphrase Albert &	256	& 43MB	& $46.8\%$ & $61.9\%$\\
    MiniLM-L6 &	256 & 120MB & $55.1\%$ & $60.3\%$ \\
    MPNET &	384 & 420MB & $63.1\%$ & $72.8\%$ \\ 
    DistilBERT & 512 & 420MB & $61.3\%$ & $74.4\%$\\
    FinBERT	& 512 &	438MB &	$48.8\%$ & $54.5\%$ \\
    \hline    
    \end{tabular}}
    \caption{Topic Classification Models Performance.}
    \label{tab:topic_perf}
\end{table}


\subsection{Training a Sentiment Model for ``Free"}

Acknowledging the significant role of emotions conveyed in earnings calls, we extended our training efforts to include a sentiment model leveraging ChatGPT's capability in handling opinions. In recent studies, there have been attempts to assign sentiment detection tasks to LLMs, such as FinBERT~\cite{yang2020finbert}, a specialized language model designed specifically for financial sentiment analysis and financial text classification tasks. In order to evaluate the quality of our sentiment models, we used an expert-labeled sentiment database (described in the benchmark datasets section), and our model achieved an accuracy of $78\%$ compared to FinBERT's $65\%$. Building upon our successful approach in topic modeling, we adopted a similar strategy for sentiment modeling, capitalizing on ChatGPT's proficiency in sentiment classification. Leveraging our existing knowledge distillation pipeline, we simply incorporated an additional prompt in the queries made to our primary model.

\begin{center}
\textbf{``Please also share your view on the financial statement's sentiment, categorizing it as either Negative, Neutral, or Positive. Structure your response in the format: Sentiment: [Negative/Neutral/Positive]."}
\end{center}

\begin{figure}
     \centering
     \includegraphics[width=0.45\textwidth]{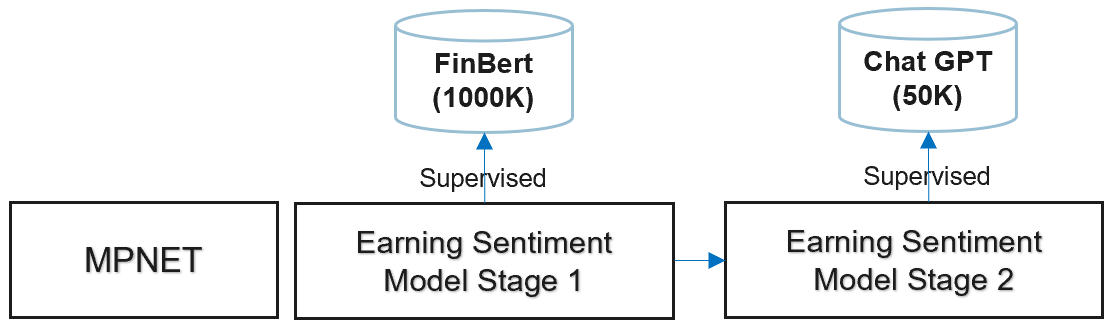}
     \caption{Sentiment Classification Model Pipeline.}
     \label{fig:sentiment_architecture}
\end{figure}

We explored two potential approaches for the sentiment model, as illustrated in Figure~\ref{fig:sentiment_architecture}. The first approach involved using only our labeled data from ChatGPT, while the second approach leveraged FinBERT as a preliminary teacher model, followed by transfer learning from the trained model using the ChatGPT dataset. Initially, training the sentiment model directly on the output of ChatGPT did not yield satisfactory results, possibly due to the limited number of data points available for sentiment training. However, the second approach yielded relatively satisfactory results, as indicated in Table~\ref{tab:sentiment_perf}. We achieved an $F_1$ score of approximately $78\%$ on our expert database, compared to $83\%$ for ChatGPT and $65\%$ for FinBERT.

\begin{table}[]
    \centering
    \resizebox{0.4\textwidth}{!}{\begin{tabular}{l|c|c|c}
    \hline
    \bf Model & \bf MPNET & \bf FinBERT	& \bf ChatGPT 3.5 \\
    \hline
    $F_1$ vs. Human & 77.8\% & 65.3\% & 83.1\% \\
    \hline
    \end{tabular}}
    \caption{Sentiment Classification Models Performance.}
    \label{tab:sentiment_perf}
\end{table}


\subsubsection{Benchmark Datasets}

To evaluate the performance of our models, we collaborated with three financial experts who tagged the topic and sentiment of 1,000 data samples (referred to as ``vs. Human" in Tables~\ref{tab:topic_perf} and \ref{tab:sentiment_perf}). Each sentence was carefully reviewed to ensure the accuracy of the assigned sentiment, and a secondary round of tagging was conducted for instances where there were discrepancies in the initial tagging.
For the topic models specifically, we set aside a separate set of data tagged by ChatGPT (referred to as ``vs. Teacher" in Table~\ref{tab:topic_perf}), which accounted for $20\%$ of our scored sample. This reserved dataset allowed us to assess the effectiveness of our supervised topic modeling approach.

\subsubsection{Computational Details}
Throughout the entire topic modeling training process, we used a hardware setup consisting of a dual Intel Xeon GOLD 3.30 Ghz processor with 280 GB of RAM, along with a RTX 4090 24GB graphics card. Each experiment took approximately 2-3 hours to complete, depending on the hyperparameter choices.
 
\section{Applications in Finance}

There are interesting applications in Finance using the models developed in the previous section. The next examples are based on historical data from the S\&P 1500 index from 2010 to 2023.

\subsection{Correlation between Topic Propensity, Sentiment Score and Sector Neutral Returns}

Let $c_i$ be the company associated with the $i$th earnings call transcript $d_i$. Each company is also associated with its GICS sector, a succinct classification of the main industry activities. For any company $c_i$, let $R_{c_i}(t+1, t+h)$ be the total returns generated by the company in the interval from time $t+1$ to $t+h$, $h>1$, including dividends. Let $\mathrm{RS}(t+1, t+h)$ be the observed capital-weighted sector returns in the same timeframe. The variable of interest is the sector neutral return (RN) observed between the time $t+1$ and $t+h$, that is, 
$$ \mathrm{RN}_{c_i}(t+1,t+h) = R_{c_i}(t+1,t+h) - \mathrm{RS}(t+1,t+h).$$

We want to evaluate statistics related to the topic model's outputs. For each document $d_i$, let $p_{ik}$ be the propensity of the $k$th topic, 
$$p_{ik} =  (1/ J_i) \sum_{j=1}^{J_i} 1_{\{T(s_{ij}) = k\}},$$ 
where $1_{\{a=b\}} = 1$ if $a=b$ and $0$ otherwise; $T(s_{ij})$ is the predicted topic of the $j$th sentence and $J_i$ is the total number of sentences in document $d_i$. This metric captures the proportion of the earnings call (in terms of number of sentences) that was allocated to the $k$th topic. Note that this quantity can either be captured by the actual prediction of the topic or by the likelihood computed by the supervised topic framework. This score is then grouped on a monthly basis creating a value for each company that had an earnings call on that given month.

Now, let $S(s_{ij}) \in \{0,1\}^3 $ be the sentiment score output where $S(s_{ij})_1$ encodes the negativity propensity, $S(s_{ij})_2$ the neutrality and $S(s_{ij})_3$ the positivity. We then define the overall sentiment score as 
$$ S(i)_k =  (1/ J_i) \sum_{j=1}^{J_i} \{S(s_{ij})_3 -S(s_{ij})_1\},$$ 
where $S(i)_k$ is only defined if $p_{ik} > 0$. This metric is capturing the ``net" positivity contained in the earnings call towards the $k$th topic.

\begin{figure}[!ht]
    \centering
     \includegraphics[width=0.23\textwidth]{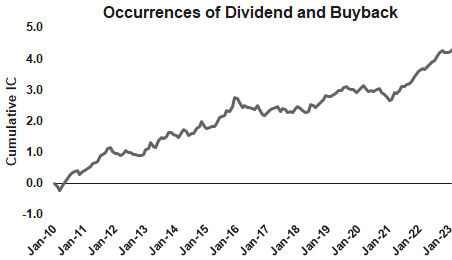}
     \includegraphics[width=0.23\textwidth]{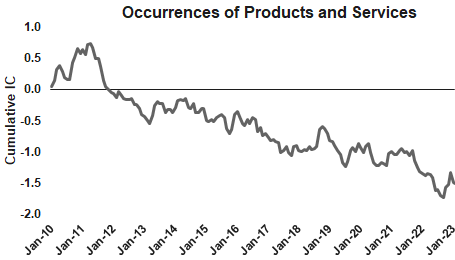}
     \includegraphics[width=0.23\textwidth]{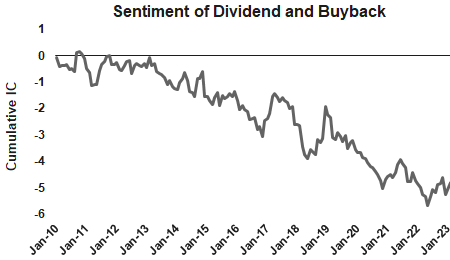}
     \includegraphics[width=0.23\textwidth]{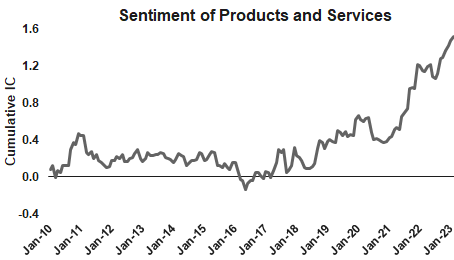}
     \caption{Cumulative IC Trends with respect to propensity and sentiment for Dividend \& Buyback (left panels) and Products \& Services (right panels).}
     \label{fig:Cumulative-IC}
 \end{figure}

In our analysis, we examined the correlation between the monthly sentiment scores $S(i)_k$ and propensity scores $p_{ik}$ with the target $\mathrm{RN}_{c_i}(t+1,t+h)$. To quantify this relationship, we used the cumulative Information Coefficient (IC), a key metric in quantitative finance that measures the predictive power of a model over time. We observed significant values of cumulative IC for certain topics identified by our model. Interestingly, the significance of sentiment, as shown in Figure~\ref{fig:Cumulative-IC}, in terms of its correlation with future performance, varied by topic and did not always align with the degree of positivity of the statements. For example, the propensity of Dividend \& Buyback is positively correlated with sector outperfomance but the sentiment is negatively correlated. In that case it's important to note that on average, the sentiment conveyed by Dividend \& Buyback is highly positive, therefore the negative correlation could be a consequence of overly positive promises that made investors second guess their assumptions. These findings indicate that effectively differentiating between topics is essential for accurate sentiment analysis of earnings calls, as this differentiation seems to play a key role in understanding their connection with future performance.

\subsection{Sentiment Gap between Sales and Earnings}

While forecasting is a fundamental aspect of quantitative investing, it's equally vital to understand current industry patterns, as this enhances confidence in and support for a particular investment strategy. Revenue and earnings are critical metrics to monitor when evaluating a company's discounted value. These two financial indicators are somewhat correlated, as earnings are derived from sales, adjusted for debt, taxes, and margins. It's common for a company to face challenges in one area while excelling in another. Changes in the sentiment expressed by company executives on these topics can offer valuable insights into an industry or a particular company, allowing investors to reevaluate their investment strategies. In our analysis of the topic model, we found that using propensity scores as filters proved to be beneficial. This technique enables us to tailor our corpus, focusing exclusively on sentences pertinent to certain financial goals. 

\begin{table}[]
\resizebox{0.45\textwidth}{!}{\begin{tabular}{l|ll|ll}
\hline
                & \multicolumn{2}{c|}{Earnings} & \multicolumn{2}{c}{Revenue} \\
                \hline
   Filter             & Outlook     & Trailling     & Outlook     & Trailling     \\
                \hline
Earnings  & High        & High          & Medium      & Medium        \\
Revenue  & Medium      & Medium        & High        & High          \\
Guidance & High        & Low           & High        & Low           \\
Others    & Low         & Low           & Low         & Low \\
\hline
\end{tabular}}
\caption{Filter intensity for earnings and revenue sentiments trends.}
\label{tab:filtering_rev}
\end{table}

\begin{figure}[!ht]
     \centering
     \includegraphics[width=0.45\textwidth]{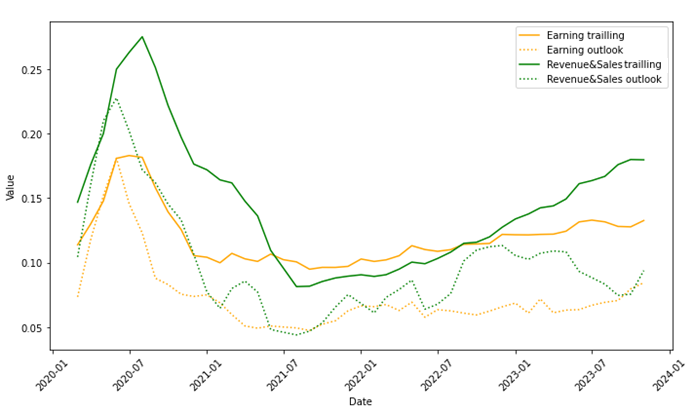}
     \caption{Trends in negativity value: earning, sales with or without outlook.}
     \label{fig:revenue-earning}
\end{figure}

\begin{figure}[!ht]
     \centering
     \includegraphics[width=0.47\textwidth]{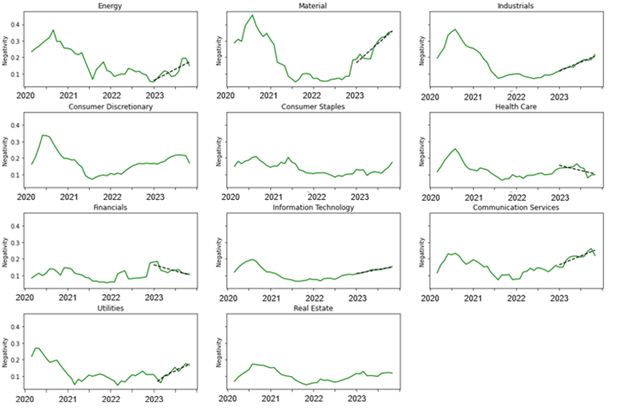}
     \caption{Negativity Trends in Revenue per GICS Sectors.}
     \label{fig:gics-revenue}
\end{figure}

Table~\ref{tab:filtering_rev} describes our filtering based on topic propensity. This creates a refined corpus primarily comprising a content relevant to examining the disparities in sentiment regarding earnings and revenue, and contrasting these with perspectives on the company's future related to these two areas. The topic \textit{``Others"} was used as a way to remove sentences that are too generic even if they are slightly related to the topic. For instance, the sentence \textit{``Jon is now going to cover our Guidance about earning"} is actually related to outlook earning but does not contain any information on what to expect from it. In order to validate the quality of our filtering, we have sampled 200 sentences for each of the areas of interest (Earnings-Revenue/ Outlook-Trailling) and have had experts classified whether the sentence was meaningfully related to the topic of interest. The sample displayed above 90\% accuracy. With that in mind, we looked at the statistics of the sentiment as displayed in Figure~\ref{fig:revenue-earning}. 
The analysis shows that the overall sentiment in regards to the revenue, within stocks composing the S\&P1500 index has significantly declined, whereas the impact on the earnings was relatively weaker.
Notably, the earnings perspective shows a consistent, less negative outlook, indicated by a nearly parallel trend between outlook and trailing data. However, recent years have seen a widening gap between outlook and trailing data, indicating increased uncertainty. Expanding this market view by sector, as illustrated in Figure~\ref{fig:gics-revenue} offers investors valuable insights into the general opinions regarding these aspects.

\section{Conclusion}


Our research introduces a new methodology for analyzing earnings calls using a knowledge distillation framework which is better adapted for use on resource-constrained devices. This method has shown potential to identify signals related to stock movements and to provide deeper insights into the content of earnings calls. There are numerous opportunities for enhancing this pipeline, especially in tailoring the approach to effectively handle sentences with multiple topics, potentially using a different loss function. Additionally, the method could leverage the physical proximity of sentences to refine topic identification (for instance, guidance statements are often found to be close to each other). We also plan to develop a mechanism for interactive dialogue with the teacher model to assess the feasibility of dividing existing topics, thereby allowing the topic model to adapt to changing market conditions.

\bibliography{bibliography}










\end{document}